\documentclass[runningheads]{llncs}
\usepackage[T1]{fontenc}
\usepackage{graphicx}
\usepackage[numbers]{natbib}

\usepackage{xfrac}
\usepackage{amsmath}
\usepackage{amssymb}
\usepackage{booktabs}
\usepackage[table]{xcolor}
\usepackage{orcidlink}

\begin{document}
\title{Variational Inference Optimized Using the Curved Geometry of Coupled Free Energy}
\titlerunning{VI using Coupled Free Energy}

\author{
Kenric P. Nelson\inst{1} \and
Igor Oliveira\inst{2} \and
Amenah Al-Najafi\inst{3} \and
Fode Zhang\inst{4} \and
Hon Keung Tony Ng\inst{5}
}

\authorrunning{K. Nelson et al.}

\institute{Photrek, LLC, 56 Burnham St Unit 1, Watertown, MA, USA \email{kenric.nelson@photrek.io} \and
Photrek, LLC, Recife, PE \and 
Department of Mathematics, University of Kufa, 299G, Najaf, Iraq
 \and
Center of Statistical Research, School of Statistics, Southwestern University of Finance and Economics, Chengdu, Sichuan 611130, PR China \and
Department of Mathematical Sciences, Bentley University Waltham, MA, USA}
\maketitle              
\begin{abstract}
We introduce an optimization framework for variational inference based on the coupled free energy, extending variational inference techniques to account for the curved geometry of the coupled exponential family. This family includes important heavy-tailed distributions such as the generalized Pareto and the Student's t. By leveraging the coupled free energy, which is equal to the coupled evidence lower bound (ELBO) of the inverted probabilities, we improve the accuracy and robustness of the learned model. The coupled generalization of Fisher Information metric and the affine connection. The method is applied to the design of a coupled variational autoencoder (CVAE).  By using the coupling for both the distributions and cost functions, the reconstruction metric is derived to still be the mean-square average loss with modified constants. The novelty comes from sampling the heavy-tailed latent distribution with its associated coupled probability, which has faster decaying tails. The result is the ability to train a model robust against severe outliers, while assuring that the training process is stable. The Wasserstein-2 or Fréchet Inception Distance of the reconstructed CelebA images shows the CVAE has a 3\% improvement over the VAE after 5 epochs of training.

\end{abstract}
\section{Introduction}
Variational Inference (VI) \citep{blei_variational_2017} is a foundational methodology in artificial intelligence, in which Bayesian learning is used to develop statistical models of complex datasets. VI could play a significant role in strengthening the capabilities of artificial general intelligence (AGI), given its relation to the neural hypothesis of predictive coding \citep{friston_predictive_2009}, in which the forecasts of higher layer neurons are compared with sensory data from lower layer neurons. Important VI algorithms include Stochastic Variational Inference \citep{ketkar_stochastic_2017, amari_backpropagation_1993}, which uses minibatches to increase the efficiency of Monte Carlo sampling; Stein Variational Gradient Descent \citep{liu_stein_2016}, which uses a kernel function for non-parametric models; and Normalizing Flows \citep{rezende_variational_2015, papamakarios_normalizing_2021}, which use invertible functions to compose a series of transformations for more complex models. We will demonstrate a generalization of VI using the Variational Autoencoder \citep{kingma_introduction_2019}, which splits the learning into a latent probability model and deterministic encoder/decoder networks. The generalization will utilize the methods of nonextensive statistical mechanics to model nonlinear dependencies and increase robustness.

VI seeks to minimize the Kullback-Liebler divergence between an approximate model $q(z)$ and an intractable posterior distribution $p(z|x)$ given a dataset $x$. Due to the complexity of the marginal distribution $p(x)$, this divergence cannot be computed directly. Instead an Evidence Lower Bound, which is the negative of the statistical Free Energy, is used for the optimization. The Free Energy consists of two components, a regularization term that measures the divergence between the latent posterior $p(z|x)$ and its prior $p(z)$, and a reconstruction accuracy that measures the log-likelihood (or cross-entropy) of the generated dataset given the original dataset. An important problem in AGI research is robust learning of complex models \cite{figurnov_robust_2016, akrami_robust_2022,li_federated_2025}. This is an important part of generalized intelligence in which models need to be robust against either training with limited datasets or operational use cases beyond the training set. Towards that end, many generalizations of the VAE have been proposed which modify the Free Energy function. Examples include use of the Wasserstein metric \citep{tolstikhin_wasserstein_2018}, adjusting the relative weight of the regularization and reconstruction loss \citep{higgins_beta-vae_2017,burgess_understanding_2018}, and utilizing generalized divergence functions \citep{kobayashis_q-vae_2020}.

Here we examine a novel design for the Coupled VAE, which uses the Coupled Exponential Family and its associated coupled entropy function, to improve the robustness and accuracy. The Coupled VAE enables the learning of heavy-tailed models (or possibly compact-support models, though that domain is not reviewed here). The core innovation is the ability to learn models assuming the presence of extreme outliers while being able to complete the training with samples from a distribution guaranteed to have a finite variance. In Section 2, we introduce the information geometry of the coupled exponential family, whose heavy-tailed (or compact-support) properties induces a curved manifold. Section 3 defines the Coupled Free Energy, including the process in which the heavy-tailed latent distribution is sampled using a modification that assures faster decaying tails. This process enables training of robust models while assuring that the training process is stable. Section 4 reviews the performance of the Coupled VAE with the Celeb-A dataset. In the conclusion, Section 5, we describe future directions for this research.

\section{The Coupled Exponential Family and Its Information Geometry}

The coupled exponential family provides a natural generalization of the classical exponential family of distributions by incorporating a measure of the degree of nonlinear coupling, $\kappa$, that modulates the long-range correlations between variates and the shape of the asymptotic tail decay. This coupling extends traditional statistical manifolds to a curved geometric setting \cite{saha_geometric_2020, frank_geometric_2021, frank_geometric_2022}, where the Fisher information metric is generalized, resulting in a non-Euclidean structure. The coupled exponential family is heavy-tailed for $\kappa>0$, exponential for $\kappa=0$, and compact-support for $-\sfrac{1}{d}<\kappa<0$, where $d$ is the dimension of the distribution. Here we evaluate the heavy-tailed domain, which is useful in assuring that AGI models are robust against outliers, however, there are also important use cases for the compact-support domain for decisive algorithms.

The standard exponential family \citep{amari_information_2016} of distributions is defined by the probability density function (PDF):
\begin{equation}
    p(\boldsymbol{x}; \boldsymbol\theta) = \frac{h(\boldsymbol{x})}{Z(\eta(\boldsymbol\theta))} \exp\left( \eta(\boldsymbol\theta) \cdot T(\boldsymbol{x}) \right),
\end{equation}
where $\boldsymbol\theta$ is the natural parameter, $T(\boldsymbol{x})$ is the sufficient statistic, $h(\boldsymbol{x})$ is a non-negative function and $Z(\eta(\boldsymbol\theta))$ is the partition function ensuring normalization. 
 
In the coupled exponential family, we introduce a deformation via a coupled logarithm and exponential transformation, defining a PDF as:
\begin{equation}\label{eq2}
p_\kappa(\boldsymbol{x}; \boldsymbol\theta) = \frac{h(\boldsymbol{x})}{Z_\kappa(\eta(\boldsymbol\theta))} \exp_\kappa^{-\frac{1+d\kappa}{\alpha}} \left( \eta(\boldsymbol\theta) \cdot T(\boldsymbol{x}) \right), 
\end{equation}
where $\exp_\kappa(x)=\left(1+\kappa x\right)^\frac{1}{\kappa}$ and its inverse is $\ln_\kappa(x)=\frac{1}{\kappa}\left(x^\kappa-1\right)$ is the coupled generalizations of the exponential and logarithm functions, and $T(\boldsymbol{x})$ is a known function, $Z_\kappa(\eta(\boldsymbol\theta))$ denotes the coupled normalizing function, $\alpha$ is a parameter to control the shape of the distribution near the location of the distribution, and $d$ is the dimension of the distribution. The exponent $-\frac{1+d\kappa}{\alpha}$ derives from the $d$ derivatives of the cdf given a survival function that starts with $\exp_\kappa(x)$. While related to the $q$-exponential family \citep{amari_geometry_2011, zhang_information_2018}, there are important distinctions. This definition ensures that each parameter is a measure of a clear, physical property of complex systems.  The functions $h$ and $Z$ are defined outside of the coupled exponential in order to keep their roles consistent with important members of the family. When brought into the coupled exponential function the coupled sum, $\exp{A}\exp{B}=\exp(A \oplus_\kappa B)$, where $A \oplus_\kappa B\equiv A+B+\kappa A B$, must be used
\begin{align}\label{11}
   \exp_\kappa^{-\frac{1+d\kappa}{\alpha}}\left(\eta(\boldsymbol\theta) \cdot T(\boldsymbol{x}) \oplus_\kappa \ln_\kappa \left(\frac{h(\boldsymbol{x})}{Z_\kappa(\eta(\boldsymbol\theta))}
\right)^{-\frac{\alpha}{1+d\kappa}}
\right).
\end{align}

Two important members of the coupled exponential family are the generalized Pareto distribution (GPD) ($\alpha=1$, coupled exponential) and the Student's t ($\alpha=2$, coupled Gaussian). For these two distributions $h(\boldsymbol{x})=1$ if natural parameters are unknown. The statistical challenge of heavy-tailed distributions is that the higher moments are undefined or diverge. For the GPD and Student's t, the $m$-th moment, $E[x^m]$, diverges if $\kappa \geq \frac{1}{m}$. Nonextensive statistical mechanics \citep{Tsallis2009a, abe_nonextensive_2001} has developed a set of modified moments based on raising the distribution by a power and renormalizing. The coupled probability \citep{nelson_average_2017} raises the distribution to a power $q=1 + \frac{\alpha\kappa}{1 + d\kappa}$, which is the Tsallis index and the fractional number of independent variates in the same state,
\begin{equation}
    P^{\left(1+\frac{\alpha\kappa}{1 + d\kappa}\right)}(x) \equiv \frac{\left(p(x)\right)^{1 + \frac{\alpha\kappa}{1 + d\kappa}}}{\int_X \left(p(x)\right)^{1 + \frac{\alpha\kappa}{1 + d\kappa}} \,dx}.
\end{equation}
The $\alpha$ parameter drops out when the coupled probability is used to compute coupled moments,
\begin{equation}
    \mathbb{E}_\kappa[x^m] \equiv \int_X x^mP^{\left(1+\frac{m\kappa}{1 + d\kappa}\right)}(x) \,dx.
\end{equation}

Let $\mathcal{S}$ be the set of coupled exponential distributions, that is
 \begin{eqnarray*}
\mathcal{S}=\left\{p_\kappa(\boldsymbol{x}; \boldsymbol\theta)\bigg|\int_{\boldsymbol{X}}p_\kappa(\boldsymbol{x}; \boldsymbol\theta) d\boldsymbol{x}=1, p_\kappa(\boldsymbol{x}; \boldsymbol\theta)\geq0, \boldsymbol\theta\in\Theta,\boldsymbol{x}\in \boldsymbol{X}\right\}.
\end{eqnarray*}
Then, the set $\mathcal{S}$ can be regarded as a statistical manifold with the parameter vector $\boldsymbol\theta$ playing the role of the coordinate system. Information geometry provides a powerful mathematical framework for analyzing and optimizing machine learning models by viewing probability distributions as points on a curved manifold. The Fisher information metric acts as a natural Riemannian metric on statistical manifolds. The Fisher information metric can be used in natural gradient descent, where it replaces the Euclidean metric used in standard gradient descent, which leads to faster convergence and improved stability, particularly in deep learning, reinforcement learning, and variational inference. Although, the performance results reported here do not include the generalized Fisher gradient, we provide the derivations in preparation for future research.

The affine connection is a tool from differential geometry that enables the definition of directional derivatives and the notion of parallel transport on curved spaces such as statistical manifolds \citep{amari_information_2010}. The affine connections offer deep insight into the structure and learning dynamics of probabilistic models, which allow for the comparison of vectors in different tangent spaces and the construction of geodesics. The following lemma gives the Fisher information metric and the affine connections of the coupled exponential family. 
\begin{lemma}\label{lemma}
 Let $\boldsymbol{\zeta} = (\boldsymbol{\theta}, \boldsymbol{x}, d, \alpha)$ and $R(\boldsymbol{\zeta}) = h(\boldsymbol{x})^{-r} \cdot Z_\kappa(\eta(\boldsymbol\theta))^{r}$ with $r = \frac{\alpha\kappa}{1 + d\kappa}$. The Fisher metric tensors and the affine connection of the coupled exponential family can be given as follows, respectively, 
 \begin{eqnarray*}
&&g_{ij} =  
 \frac{1}{\alpha} 
\mathbb{E}_{\boldsymbol{X}}\left[B_1+B_2\right] 
\\
&&\Gamma_{ijk} 
= \frac{1}{\alpha^2} \cdot \mathbb{E}_{\boldsymbol{X}}\left[ (A_1 + A_2) \cdot (B_1 + B_2) \right]
\end{eqnarray*}
where
\begin{align*}
    A_1 &:= \left[T(\boldsymbol{x})\right] \left( \frac{\partial R(\boldsymbol{\zeta})}{\partial \theta_j} + \frac{\partial R(\boldsymbol{\zeta})}{\partial \theta_i} \right),\\ 
A_2 &:= \left( \frac{1}{\kappa} + T(\boldsymbol{x}) \cdot \boldsymbol{\theta} \right) \cdot \frac{\partial^2 R(\boldsymbol{\zeta})}{\partial \theta_j \partial \theta_i},\\
B_1 &:= T(\boldsymbol{x}) \cdot R(\boldsymbol{\zeta}),\\
B_2 &:= \left( \frac{1}{\kappa} + T(\boldsymbol{x}) \cdot \boldsymbol{\theta} \right) \cdot \frac{\partial R(\boldsymbol{\zeta})}{\partial \theta_k}.
\end{align*}
\end{lemma}

The proof of Lemma \ref{lemma} is in the attachment \citep{amenah_coupled_2025}.

This modification results in a curved manifold whose geometry deviates from the standard dually-flat structure of classical exponential families. The curvature of the coupled exponential family influences the optimization landscape in variational inference. Specifically, the variational coupled free energy, or equivalently, the coupled ELBO of the inverse probabilities, now operates over a non-trivial geometric structure, modifying the trajectories of inference and learning dynamics. This formulation suggests that variational inference architectures, including predictive coding, leveraging coupled geometries may achieve more robust uncertainty quantification and improved representation of complex data distributions.

\section{Coupled Free Energy and Variational Inference}
The coupled exponential family induces a generalization of information theory. The foundations of this are described by the literature on nonextensive statistical mechanics; however, the resulting coupled entropy $(H_\kappa)$ is equal to the Normalized Tsallis entropy divided by $1+\kappa$.  The coupled entropy is defined as:
\begin{eqnarray*} 
H_\kappa \left(\textbf{p}; \alpha, d \right) & = & \frac{1}{\alpha} \sum_{i=1}^{N} P_i^{(1+\frac{\alpha\kappa}{1+d\kappa})} \ln_\kappa \left(p_i^{\frac{-\alpha}{1+d\kappa}} \right)\\ 
& = & \frac{1}{\alpha} \ln_\kappa \left(\sum_{i=1}^{N} p_i^{1 + \frac{\alpha \kappa}{1+d\kappa}} \right)^{\frac{1 + d \kappa}{\alpha \kappa}}.
\end{eqnarray*}
The structure of the coupled entropy is such that as the coupling increases the coupled logarithm term $\ln_\kappa \left(p_i^{\frac{-\alpha}{1+d\kappa}} \right)\rightarrow\infty$ faster as $p_i\rightarrow0$. At the same time for larger coupling, the coupled probability term, $P_i^{(1+\frac{\alpha\kappa}{1+d\kappa})}$, defines a distribution with faster decaying tails. The coupled probability defines a subsampling of the original distribution constricted to  $q=(1+\frac{\alpha\kappa}{1+d\kappa})$ random variables in the same state.  This subsampling has the effect of increasing the decay of the tail and thus reducing the outlier samples.  There is thus a balance between a increasing the cost of outliers while averaging over a distribution with fewer outliers. We'll show that for machine learning this provides a methodology in which systems can be trained to be robust against outliers while assuring that Monte Carlo sampling does not result in instability during the training. 

From the coupled information theory, we derive the coupled free energy (CFE) function which provides a non-euclidean metric.  In the standard VAE, the learning objective based on the negative ELBO or FE is based on the euclidean geometry of the Gaussian distribution, the Kullback–Leibler (KL) divergence, and the log-likelihood \citep{kingma_introduction_2019}. However, many real life data distributions exist on manifolds which have non-Euclidean geometric structures \citep{amari_information_2016}. The non-euclidean manifold can be used to improve the accuracy and robustness of the learned model. Improvements in accuracy are also possible, since the multivariate heavy-tailed distribution captures nonlinear correlations in the dataset. The robustness derives from the fact that CFE increases the penalties in the tails of the distribution. 

Both the divergence and log-likelihood components of the FE are generalized using the coupled algebra. Training a coupled VAE with the CFE was first reported by Cao, et. al \citep{cao_coupled_2022}; however, in that design samples were drawn directly from a Gaussian multivariate distribution. Here we report on the performance with a) a coupled Gaussian latent distribution, b) a CFE function with a matching coupling parameter, and c) a training process that uses samples drawn from the coupled probability of the coupled Gaussian  distribution $( Q^{\left(\frac{2\kappa}{1+d\kappa}\right)})$, allowing controlled tail behavior during optimization.  This approach makes full utilization of the coupled information theory and results in two important innovations. First, the coupled log-likelihood still simplifies to the mean-square average between the reconstructed and original data with only a modification in the constants. Second, even extreme models assuming a delta-function like distribution with excessive outliers can be trained, since the training samples are guaranteed to be drawn from a less extreme distribution with a finite variance.

Here, we denote the coupled free energy (CFE) by $\mathcal{F}_\kappa$, which corresponds to the Evidence Lower Bound (ELBO) applied to the inverse of the probabilities. In the following, we extend the VAE to the case where both the prior and posterior are coupled Gaussian distributions, for which $\alpha=2$. Let $A_q$ and $A_p$ denote the normalization terms of the coupled posterior and prior distributions, respectively, where 
$
A_q = \left(\ln_\kappa \left(\frac{1}{Z_q} \right)\right)^{- \frac{2}{1 + d \kappa}}$ and  $A_p = \left[\ln_\kappa \left(\frac{1}{Z_p} \right)\right]^{- \frac{2}{1 + d \kappa}}
$.

\begin{theorem}[Coupled Free Energy for Multivariate Coupled  Gaussian]
Let $\boldsymbol{x} \in \mathbb{R}^d$ and $\mathbf{z} \in \mathbb{R}^n$ be random vectors, and suppose that the posterior $q(\mathbf{z} \mid \boldsymbol{x})$ and the prior $p(\mathbf{z})$ are both multivariate coupled Gaussian with their joint PDFs  defined as:
\small
\begin{equation*}
f\left(\boldsymbol{x}; \boldsymbol{\mu}, \boldsymbol{\Sigma}, \kappa \right) \equiv 
\begin{cases}
\frac{1}{Z(\boldsymbol{\Sigma}, \kappa)}
\left(1 + \kappa \left|(\boldsymbol{x} - \boldsymbol{\mu})^{\intercal} \boldsymbol{\Sigma}^{-1} (\boldsymbol{x} - \boldsymbol{\mu})\right|\right)_+^{-\frac{1 + d\kappa}{2 \kappa}}, & \kappa \ne 0, \kappa>-\sfrac{1}{d}; \\
\frac{1}{(2\pi)^{d/2} |\boldsymbol{\Sigma}|^{1/2}} \exp\left(-\frac{1}{2}(\boldsymbol{x} - \boldsymbol{\mu})^\top \boldsymbol{\Sigma}^{-1}(\boldsymbol{x} - \boldsymbol{\mu})\right), & \kappa = 0.
\end{cases} \\
\end{equation*}
Given a Coupled Variational Autoencoder (CVAE) with $\kappa \ne0$, the Coupled Free Energy (CFE) takes the following form with a coupled divergence term plus a reconstruction loss term:

\begin{center}
\scalebox{0.75}{$
\begin{aligned}
\mathcal{F}_{\theta,\phi,\kappa}(\boldsymbol{x}) &= 
\frac{1}{2}\mathbb{E}_{\mathbf{z} \sim Q_\phi^{\left(\frac{2\kappa}{1+d\kappa}\right)}(\mathbf{z}\mid \boldsymbol{x})}
\left[\ln_{\kappa}\left(p_\theta(\mathbf{z})^{\frac{-2}{1+d\kappa}}\right) - 
\ln_{\kappa}\left(q_\phi(\mathbf{z}\mid \boldsymbol{x})^{\frac{-2}{1+d\kappa}}\right) 
\right]\,d\mathbf{z} \\
&\quad +\frac{1}{2}\mathbb{E}_{\mathbf{z} \sim Q_\phi^{\left(\frac{2\kappa}{1+d\kappa}\right)}(\mathbf{z}\mid \boldsymbol{x})}
\left[
\ln_{\kappa}\left(
\exp_\kappa^{\frac{-(1+d\kappa)}{2}} 
\left(
(\boldsymbol{x} - \bar{\boldsymbol{x}}_{x|z})^\top \mathbf{\Sigma}_{x|z}^{-1} (\boldsymbol{x} - \bar{\boldsymbol{x}}_{x|z})
\oplus_\kappa 
\ln_\kappa\left(\frac{1}{Z_q}\right)^{\left( \frac{-2}{1+d\kappa} \right)}
\right)\right)^\frac{-2}{1+d\kappa}
\right]
\end{aligned}
$}
\end{center}

\vspace{1em}
The CFE simplifies to:
\begin{center}
\scalebox{0.85}{$
\begin{aligned}
\mathcal{F}_{\theta,\phi,\kappa}(\boldsymbol{x}) &= 
-\frac{d\left(1+\kappa A_q\right)}{2} + \frac{1+\kappa A_p}{2}
\left((\boldsymbol{\mu}_p - \boldsymbol{\mu}_q)^T \mathbf{\Sigma}_p^{-1}(\boldsymbol{\mu}_p - \boldsymbol{\mu}_q) 
+ \operatorname{tr}\left(\mathbf{\Sigma}_p^{-1} \mathbf{\Sigma}_q\right)\right) \\
&\quad - \frac{A_q}{2} + \frac{A_p}{2} 
- \mathbb{E}_{\mathbf{z} \sim Q_\phi^{\left(\frac{2\kappa}{1+d\kappa}\right)}(\mathbf{z} \mid \boldsymbol{x})}
\left[ 
\frac{1}{2}
\left( 
(\boldsymbol{x} - \bar{\boldsymbol{x}}_{x|z})^\top \mathbf{\Sigma}_{x|z}^{-1} (\boldsymbol{x} - \bar{\boldsymbol{x}}_{x|z}) 
\oplus_\kappa A_{x|z} 
\right)
\right]
\end{aligned}
$}
\end{center}

\end{theorem}
This result provides a closed-form expression for the coupled Free Energy in terms of the coupled divergence and reconstruction loss. We emphasize that the expectation in the CFE expression is taken with respect to a transformed sampling distribution, denoted as $( Q^{\left(\frac{2\kappa}{1+d\kappa}\right)}(\mathbf{z} \mid \mathbf{x}) )$, rather than the original posterior density $( q(\mathbf{z} \mid \mathbf{x}))$. This transformation modifies both the shape and the scale of the distribution, effectively compressing the tails while preserving the location. This transformed distribution corresponds to a coupled Gaussian with an adjusted coupling parameter and precision matrix. The transformed coupling ($ \kappa_Q $) and scale matrix ($ \mathbf{\Sigma}_Q $) are given by:
\begin{align}
\kappa_Q &= \frac{\kappa_q}{1 + 2\kappa_q}, \\
|\mathbf{\Sigma}_Q|^{-1} &= (1 + 2\kappa_q) |\mathbf{\Sigma}_q|^{-1}.
\end{align}
These complementary modifications are such that the exponent of the distribution is modified while the multiplicative term is unchanged:
\[
\kappa_q (\mathbf{z} - \boldsymbol{\mu}_q)^\top \mathbf{\Sigma}_q^{-1} (\mathbf{z} - \boldsymbol{\mu}_q)
= \kappa_Q (\mathbf{z} - \boldsymbol{\mu}_q)^\top \mathbf{\Sigma}_Q^{-1} (\mathbf{z} - \boldsymbol{\mu}_q).
\]
Even if the latent distribution model is a delta function, $\kappa\rightarrow\infty$, the distribution used for the training will have a well-defined mean and finite variance, $\kappa\rightarrow\sfrac{1}{2}$. The full derivation of the coupled Free Energy expression is available at \cite{amenah_coupled_2025}. 

\section{Experimental Validation}
In this section, we show the initial results of the CVAE performance. To train the model, we consider the CelebA dataset (CelebFaces Attributes Dataset), a widely used dataset for computer vision tasks involving human faces. This data set contains $202,599$ images of celebrity faces with high variability and visual richness, making it an ideal benchmark for learning complex data distributions and testing the performance of generative models. For testing purposes, we use an image size of $128\times128$ (original images are $178\times178$), a batch size of $64$ images, a learning rate of $ 5\times10^ {-4}$, and a latent dimension of $d =10$.
\vspace{-5pt}
\begin{figure}[ht!]
     \centering
		\includegraphics[width= \linewidth]{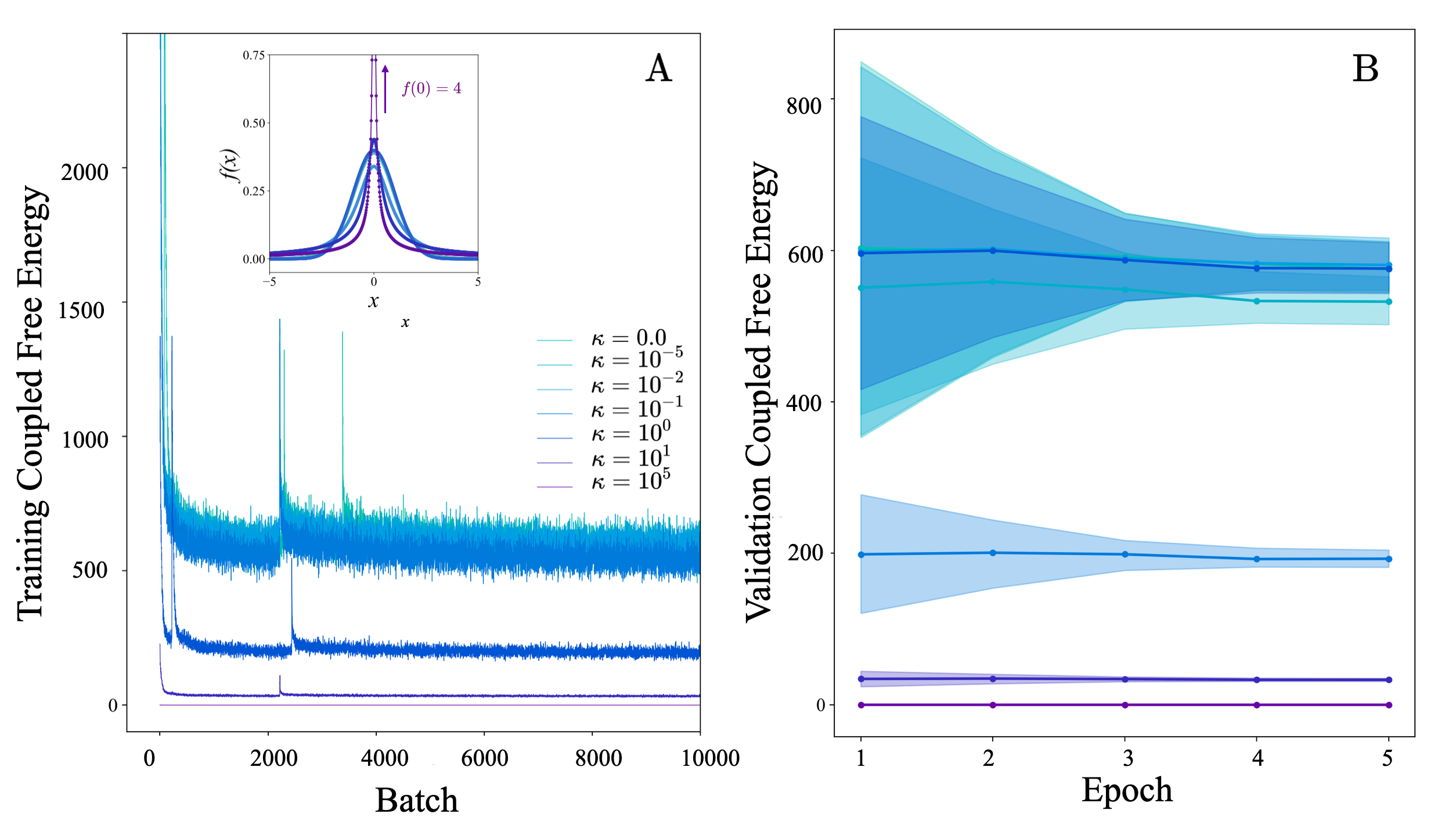} 
            \caption{Comparison of the Coupled Free Energy across different values of the coupling parameter $\kappa$ for both (A) training and (B) validation phases using diagonal covariance. The models are trained and validated using the coupled probability distribution $Q$. The inset in panel (A) shows the shape of the coupled Gaussian distributions for each $\kappa$, displaying the algorithm's capability to train extreme heavy-tailed distributions. The error bars in panel (B) represent the standard deviation of the Free Energy values across each epoch over several batches.}
		\label{Fig_energy.png}
\end{figure}

The model was trained using Adam Optimizer and without a learning rate scheduler. Gradient clipping with a maximum norm of 10 was applied to prevent exploding gradients and improve stability. To ensure reproducibility and consistent weight initialization, all linear and transposed convolution layers were initialized using the Kaiming uniform method (for layers with LeakyReLU activations), while linear layers used Xavier initialization. Batch normalization layers were initialized with unit scale and zero bias, and their running statistics were reset when loading checkpoints, helping stabilize the internal covariate shift during training. The CelebA data set was randomly split into $70\%$ training, $15\% $ validation, and $15\%$ test sets, and the data was shuffled at the beginning of each epoch. The models were trained and validated using the coupled probability distribution $Q$ samples. For validation and testing, samples were drawn from the coupled Gaussian $q$ and its adjusted version $Q$ for comparison. A fixed batch of validation images generated consistent reconstructions and samples across epochs for visual assessment.

The architecture employed for the VAE and CVAE models consists of a convolutional encoder-decoder structure. The encoder is built from four convolutional layers with increasing depth (32 to 256 channels), followed by batch normalization and LeakyReLU activation functions. The encoded feature map is flattened and passed through two linear layers to produce the latent mean and log-variance vectors. The decoder mirrors the encoder structure with transposed convolutions, batch normalization, and LeakyReLU activations, culminating in a final Sigmoid layer to map outputs to the $[0, 1]$ range. 

In these initial results, we conducted experiments using PyTorch's automatic differentiation framework (autograd), which computes gradients based on the standard Euclidean geometry of the parameter space. Although this approach is sufficient for baseline comparisons and initial performance assessments, it does not account for the curved geometry induced by the coupled exponential family. Future work will incorporate the information geometric formulation introduced in this paper, using the analytically derived gradients and curvature-aware optimization based on the coupled free energy. This is expected to improve the efficiency and stability of the learning convergence by aligning the optimization trajectory with the intrinsic geometry of the coupled statistical manifold.

\begin{figure}[ht!]
 \centering
		\includegraphics[width= \linewidth]{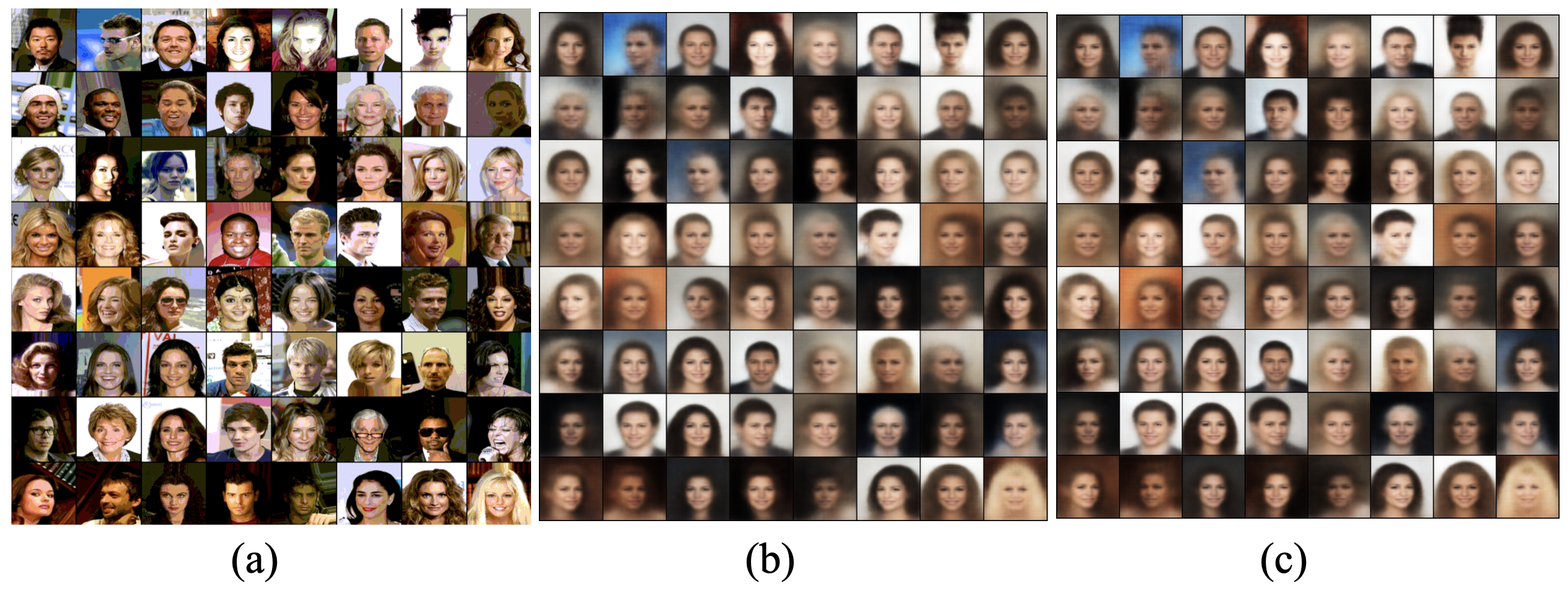} 
           \caption{Comparison of the standard linear correlation and the long-range coupling models.  (a) Original images and reconstructions with (b) Diagonal standard VAE and (c) Diagonal Coupled VAE with $\kappa = 1$. }
           \vspace{-5pt}
		\label{Fig_reconstructions.png}
\end{figure}

In Fig. \ref{Fig_energy.png} we show that the CVAE with $\kappa = 10^{-5}$ produces a stationary energy similar to the standard VAE $(\kappa=0)$, but the CVAE with $\kappa = 1, 10$ and $\kappa = 10^5$ shows significantly lower coupled free energy with both the training and validation data. Additionally, the CVAE energy curves with positive coupling are less noisy and have fewer and smaller spikes, which may indicate that the coupled model was trained more robustly. Because the coupled free energy metric varies with the coupling, these measurements do not provide a direct comparison; however, as the coupling increases, the metric increases in sensitivity and can rise above the $\kappa=0$ metric if the algorithm lacks robustness. For comparison, Fig. \ref{Fig_reconstructions.png} shows the standard and coupled VAE reconstructions, with the latter capturing higher fidelity. 
\vspace{-5pt}
We employ several widely accepted metrics to assess the quality of the generated reconstructions, including the Fréchet Inception Distance (FID) and the Kernel Inception Distance (KID). Lower FID values indicate a closer alignment between the real and generated image distributions, reflecting higher visual fidelity and diversity. KID, in contrast, avoids the Gaussian assumption by directly computing the squared Maximum Mean Discrepancy (MMD) using a polynomial kernel. To capture perceptual fidelity, we report the Learned Perceptual Image Patch Similarity (LPIPS), a metric based on deep network embeddings that aligns closely with human visual perception. Additionally, we include pixel- and structure-level similarity metrics: Multi-Scale Structural Similarity Index (MS-SSIM), Peak Signal-to-Noise Ratio (PSNR), and Structural Similarity Index (SSIM). Beyond fidelity-based metrics, we evaluate distributional alignment using the Fréchet ResNet Distance (FRD), an alternative to FID based on ResNet-derived features. 

The best-performing $\kappa$ values were $10^{-5}$,  $10^{-1}$, and  $10^{0}$, each achieving a strong balance between statistical alignment (low FID, KID, and FRD) and visual or perceptual fidelity (low LPIPS, high MS-SSIM, PSNR, and SSIM). Among these, $\kappa = 10^{-5}$ yielded the best overall equilibrium, slightly outperforming others in FRD and reconstruction accuracy, while $\kappa = 1$ achieved the highest Precision, indicating superior sample quality. The setting $\kappa = 10^{-1}$ consistently performed well across all metrics, highlighting it as a reliable and stable configuration. The baseline model with $\kappa = 0$ achieved reasonable statistical alignment but underperformed in visual fidelity. The use of the adjusted distribution $Q$ significantly stabilized the algorithm, enabling proper running even with a delta-function-like distribution, $\kappa = 10^5$. All metrics were computed over multiple reconstruction samples for each $\kappa$ setting, and the uncertainties in parenthesis reported in Table~\ref{Table metrics} represent the standard deviations.
\vspace{-5pt}
\begin{table}[h!]
\centering
\renewcommand{\arraystretch}{1.4}
\setlength{\tabcolsep}{6pt}
\resizebox{\textwidth}{!}{%
\begin{tabular}{lccccccccc}
\toprule
\rowcolor{gray!20}
\textbf{$\kappa$} & \textbf{FID} & \textbf{KID} & \textbf{LPIPS} & \textbf{MS-SSIM} & \textbf{PSNR} & \textbf{SSIM} & \textbf{FRD} & \textbf{Prec} & \textbf{Rec} \\
\midrule
$0.0$        & 15.0(9) & 0.212(3) & 0.503(4) & 0.478(7) & 14.3(2) & 0.503(6) & 105.0(6) & 0.828(2) & 0.556(7) \\
$10^{-5}$    & \textbf{\textcolor{green!50!black}{15.0(7)}} & \textbf{\textcolor{green!50!black}{0.209(2)}} & \textbf{\textcolor{green!50!black}{0.460(1)}} & \textbf{\textcolor{green!50!black}{0.596(5)}} & \textbf{\textcolor{green!50!black}{17.6(9)}} & \textbf{\textcolor{green!50!black}{0.578(6)}} & \textbf{\textcolor{green!50!black}{103.5(6)}} & 0.822(1) & \textbf{\textcolor{green!50!black}{0.569(5)}} \\
$10^{-2}$    & 15.3(6) & 0.214(2) & 0.579(4) & 0.323(4) & 9.30(4) & 0.361(4) & 106.8(6) & 0.753(1) & 0.491(5) \\
$10^{-1}$    & \textbf{\textcolor{green!50!black}{14.9(6)}} & \textbf{\textcolor{green!50!black}{0.209(2)}} & \textbf{\textcolor{green!50!black}{0.461(1)}} & \textbf{\textcolor{green!50!black}{0.595(0)}} & \textbf{\textcolor{green!50!black}{17.6(1)}} & \textbf{\textcolor{green!50!black}{0.577(0)}} & \textbf{\textcolor{green!50!black}{103.3(5)}} & \textbf{\textcolor{green!50!black}{0.838(1)}} & \textbf{\textcolor{green!50!black}{0.566(8)}} \\
$10^{0}$     & \textbf{\textcolor{green!50!black}{14.9(6)}} & 0.213(2) & \textbf{\textcolor{green!50!black}{0.461(1)}} & \textbf{\textcolor{green!50!black}{0.595(0)}} & \textbf{\textcolor{green!50!black}{17.6(2)}} & \textbf{\textcolor{green!50!black}{0.578(5)}} & \textbf{\textcolor{green!50!black}{104.3(5)}} & \textbf{\textcolor{green!50!black}{0.853(2)}} & 0.541(6) \\
$10^{1}$     & 15.0(7) & 0.209(2) & \textbf{\textcolor{green!50!black}{0.460(1)}} & \textbf{\textcolor{green!50!black}{0.594(5)}} & \textbf{\textcolor{green!50!black}{17.6(2)}} & \textbf{\textcolor{green!50!black}{0.577(3)}} & \textbf{\textcolor{green!50!black}{104.4(6)}} & 0.825(4) & \textbf{\textcolor{green!50!black}{0.588(2)}} \\
$10^{5}$     & 15.8(1) & 0.235(2) & 0.473(10) & 0.587(6) & 17.3(2) & 0.571(8) & 107.9(6) & 0.838(5) & 0.513(5) \\
\bottomrule
\end{tabular}}
\caption{Comprehensive evaluation of reconstruction quality across different values of $\kappa$ using diagonal covariance in the Coupled Variational Autoencoder (CVAE). Metrics include FID, KID, LPIPS, MS-SSIM, PSNR, SSIM, FRD, Precision, and Recall, with the best results per metric highlighted in green. Lower FID, KID, LPIPS, and FRD values indicate better perceptual and distributional similarity, while higher MS-SSIM, PSNR, SSIM, Precision, and Recall reflect improved visual fidelity and alignment. Reconstructions were generated using coupled Gaussian samples $Q$, which reduce to standard Gaussian samples when $\kappa = 0$. The results suggest that moderate coupling values (e.g., $\kappa = 10^{-1}$ and $\kappa = 10^{0}$) yield the best trade-off across most quality metrics. The number in parenthesis is the standard deviation spread of the last digit.} 
\label{Table metrics}
\vspace{-10pt}
\end{table}

\section{Conclusion}
We present a generalization of VI, using the Coupled Free Energy (Coupled Evidence Lower Bound of the inverse probabilities), which enables the training of robust, complex models using robust samples. The method utilizes the geometric curvature induced by the heavy-tailed characteristics of the coupled exponential family of distributions. The example of the VAE is used to demonstrate the improvement in the ability to learn complex datasets. The robustness of the models is improved by using a heavy-tailed latent distribution (the coupled Gaussian) and by increasing the cost of outliers via the coupled Free Energy. 

An original element of the design is that the statistical robustness of the learning is improved by drawing samples from the coupled probability (also known as an escort probability) of the latent distribution. The coupled probability, which raises the distribution to a power and renormalizes, reduces the distribution's scale and shape. The coupled probability samples are equivalent to conditioning the original distribution to subselect for samples in which $q$ of them are equal.  The effect is to filter extreme outliers from the training process, while still assuring that the learned model will be robust against those outliers. While consistent with the theory of nonextensive statistical mechanics, this is a novel mechanism in that the coupled entropy is a compromise between the Tsallis entropy and the normalized Tsallis entropy. 

Preliminary evidence of the robustness of this CVAE design is provided by the significant reduction in the CFE with $\kappa=1$ versus the FE with $\kappa=0$ measurements. While these are different metrics given the change in $\kappa$, non-robust models would be susceptible to an increase over the FE metric. A 3\% improvement in the accuracy of the reconstructions was measured using the Fréchet Inception Distance, which is equivalent to the Wasserstein-2 metric.

While the Coupled Free Energy function derivation is complex, the machine learning implementation is relatively simple for a couple of reasons.
    \begin{itemize}
        \item First, the coupled Gaussian (Student's t) is a well-established model. It was first used by Guinness Brewing to improve their beer production.
        \item Second, we establish how to train quite extreme heavy-tailed models, while drawing samples from significantly faster decaying distributions.
        \item The reconstruction likelihood is proven to be the same as the original Free Energy, a mean-square average, with just a modification of the constants.
    \end{itemize}

Our future research will include the computation of the curved gradients using the coupled affine connection to improve the speed and stability of the learning rate. This first demonstration utilized the standard Euclidean gradients of the coupled Free energy. Additional research objectives are to explore methods to determine criteria for an optimal coupling value and the limits of how high the coupling can be set. Negative coupling values can also be valuable in producing models intended to be more decisive and extreme. The coupled VAE is just one example of the many techniques in VI that can be generalized to undertake more complex AI tasks.

%
%
%
\bibliographystyle{unsrtnat}
\bibliography{MICS}
%

\end{document}